%% file: main.tex
\pdfoutput=1

\documentclass[11pt]{article}
\usepackage[table]{xcolor}

\usepackage[]{ACL2023}

\usepackage{times}
\usepackage{latexsym}

\usepackage[T1]{fontenc}

\usepackage[utf8]{inputenc}

\usepackage{microtype}

\usepackage{inconsolata}
\usepackage{booktabs}
\usepackage{multirow}
\usepackage{makecell}
\usepackage{graphicx}
\usepackage{amsthm}
\usepackage{amssymb}

\input{code}

\title{Benchmarking Retrieval-Augmented Generation for Medicine}

\author{
\textbf{Guangzhi Xiong}$^{\clubsuit\dagger}$, \textbf{Qiao Jin}$^{\heartsuit\dagger}$, \textbf{Zhiyong Lu}$^{\heartsuit\S}$, \textbf{Aidong Zhang}$^{\clubsuit\S}$ \\
$^{\clubsuit}$ \text{Univeristy of Virginia} \\
$^{\heartsuit}$ \text{National Library of Medicine, National Institutes of Health} \\
\texttt{\{hhu4zu, aidong\}@virginia.edu} \\
\texttt{\{qiao.jin, zhiyong.lu\}@nih.gov}}

\begin{document}
\maketitle
\def\thefootnote{$\dagger$}\footnotetext{Equal contribution.}
\def\thefootnote{$\S$}\footnotetext{Co-correspondence.}

\renewcommand{\thefootnote}{\arabic{footnote}} 
\setcounter{footnote}{0} 

\input{content/abstract}
\input{content/introduction}
\input{content/related_work}
\input{content/benchmark}
\input{content/method}
\input{content/results}
\input{content/discussion}
\input{content/conclusion}

\section*{Limitations}
While our study provides systematic evaluations and practical recommendations for medical RAG systems, there are several limitations that need to be acknowledged.
First, there have been novel developments in the architecture of RAG (e.g., active RAG, \citeauthor{jiang2023active}, \citeyear{jiang2023active}). However, we mainly evaluate the vanilla RAG architecture where the retrieved documents are directly prepended in the LLM context because this is the most widely implemented architecture. Evaluating new RAG system designs remains an important direction to explore.
Second, while the coverage of corpora, retrievers, and LLMs in \textsc{MedRag} is reasonably comprehensive, there are other potentially useful resources that can also be incorporated into \textsc{MedRag} in future work, such as the full-text articles from PubMed Central (PMC)\footnote{\url{https://www.ncbi.nlm.nih.gov/pmc/}} and Frequently Asked Questions (FAQs) from trustworthy sources \citep{ben2019question}. 
Third, we only evaluate the retrieval component for PubMedQA* and BioASQ-Y/N since the other three examination datasets lack labels of ground-truth supporting documents. Further research should also evaluate whether the retrieved snippets are actually helpful for the examination datasets, and explore the use of cross-encoder re-rankers to improve the retrieval performance for relevant information.
Fourth, while QA is the most commonly used task for evaluating biomedical LLMs, there are also other knowledge-intensive tasks that might benefit from \textsc{MedRag}, such as claim verification \citep{wadden2020fact,liu2024retrieval}. Following most other studies, we use the format of multi-choice questions for large-scale and automatic evaluation of medical QA. Although we restrict the retrieval phase to having no access to the choices, LLMs still need to use them as input for the final prediction. The rationales generated by \textsc{MedRag} remain to be evaluated as well. As the goal of this study is to systematically benchmark the most commonly used medical RAG settings, we leave the potential solutions of the above-mentioned limitations to future work.

\section*{Acknowledgements}
Guangzhi Xiong and Aidong Zhang are supported by NIH grant 1R01LM014012 and NSF grant 2333740.
Qiao Jin and Zhiyong Lu are supported by the NIH Intramural Research Program, National Library of Medicine.

\bibliography{main}
\bibliographystyle{main}

\clearpage
\appendix

\section*{Appendix}
\input{content/benchmark_appendix}
\input{content/method_appendix}

\input{content/appendix}

\end{document}

%% file: code.tex
\usepackage[most]{tcolorbox}
\usepackage{listings}
\usepackage{float}

\tcbset{
  aibox/.style={
    width=474.18663pt,
    top=10pt,
    colback=white,
    colframe=black,
    colbacktitle=black,
    enhanced,
    center,
    attach boxed title to top left={yshift=-0.1in,xshift=0.15in},
    boxed title style={boxrule=0pt,colframe=white,},
  }
}
\newtcolorbox{AIbox}[2][]{aibox,title=#2,#1}

%% file: content/abstract.tex
\begin{abstract}
While large language models (LLMs) have achieved state-of-the-art performance on a wide range of medical question answering (QA) tasks, they still face challenges with hallucinations and outdated knowledge. Retrieval-augmented generation (RAG) is a promising solution and has been widely adopted. However, a RAG system can involve multiple flexible components, and there is a lack of best practices regarding the optimal RAG setting for various medical purposes. To systematically evaluate such systems, we propose the Medical Information Retrieval-Augmented Generation Evaluation (\textsc{Mirage}), a first-of-its-kind benchmark including 7,663 questions from five medical QA datasets. Using \textsc{Mirage}, we conducted large-scale experiments with over 1.8 trillion prompt tokens on 41 combinations of different corpora, retrievers, and backbone LLMs through the \textsc{MedRag} toolkit introduced in this work. Overall, \textsc{MedRag} improves the accuracy of six different LLMs by up to 18\% over chain-of-thought prompting, elevating the performance of GPT-3.5 and Mixtral to GPT-4-level. Our results show that the combination of various medical corpora and retrievers achieves the best performance. In addition, we discovered a log-linear scaling property and the ``lost-in-the-middle'' effects in medical RAG. We believe our comprehensive evaluations can serve as practical guidelines for implementing RAG systems for medicine.
\end{abstract}

%% file: content/introduction.tex
\section{Introduction}
Large Language Models (LLMs) have revolutionized the way people seek information online, from searching to directly asking chatbots for answers.
Although recent studies have shown their state-of-the-art capabilities of question answering (QA) in both general and medical domains \citep{openai2023gpt4,anil2023palm,touvron2023llama2,singhal2023large,nori2023capabilities}, LLMs often generate plausible-sounding but factually incorrect responses, commonly known as hallucination \cite{ji2023survey}.
Also, the training corpora of LLMs might not include the latest knowledge, such as recent updates of clinical guidelines.
These issues can be especially dangerous in high-stakes domains such as healthcare \cite{tian2024opportunities,hersh2024search}. 

By providing LLMs with relevant documents retrieved from up-to-date and trustworthy collections, Retrieval-Augmented Generation (RAG) has the potential to address the above challenges \cite{lewis2020retrieval, gao2023retrieval}.
RAG also improves the transparency of LLMs by grounding their reasoning on the retrieved documents.
As such, RAG has already been quickly implemented in various scientific and clinical QA systems \cite{lala2023paperqa,zakka2024almanac}.
However, a complete RAG system contains several flexible modules, such as document collections (corpora), retrieval algorithms (retrievers), and backbone LLMs, but the best practices for tuning these components are still unclear, hindering their optimal adoption in medicine.

To systematically evaluate how different components in RAG affect its performance, we first compile an evaluation benchmark termed \textsc{Mirage}, representing Medical Information Retrieval-Augmented Generation Evaluation. 
\textsc{Mirage} includes 7,663 questions from five commonly used QA datasets in biomedicine. To evaluate RAG in realistic medical settings, \textsc{Mirage} focuses on the zero-shot ability in RAG systems where no demonstrations are provided. We also employ a question-only setting for the retrieval phase of RAG, as in real-world cases where no options are given. For a comprehensive comparison on \textsc{Mirage}, we provide \textsc{MedRag}, an easy-to-use toolkit that covers five corpora, four retrievers, and six LLMs including both general and domain-specific models.

Based on the \textsc{Mirage} benchmark, we systematically evaluated different \textsc{MedRag} solutions and studied the effects of each component on overall performance from a multidimensional perspective. For various \textbf{LLMs}, there is a 1\% to 18\% relative performance increase using \textsc{MedRag} compared to chain-of-thought prompting \citep{wei2022chain}. 
Notably, with \textsc{MedRag}, GPT-3.5 and Mixtral \cite{jiang2024mixtral} can achieve comparable performance to GPT-4 \cite{openai2023gpt4} on \textsc{Mirage}.
On the \textbf{corpus} dimension, we found different tasks have a preference over the retrieval corpus. While point-of-care articles and textbooks are solely helpful for examination questions, PubMed is a robust choice for all \textsc{Mirage} tasks. Our results also show that a combination of all corpora can be a more comprehensive choice. On the \textbf{retriever} dimension, BM25 \cite{robertson2009probabilistic} and the domain-specific MedCPT \cite{jin2023medcpt} retriever display superior performance on our \textsc{Mirage} benchmark. The performance can be further enhanced by combining multiple retrievers. Beyond the evaluation results on \textsc{Mirage}, we found a log-linear scaling relationship between model performance and the number of retrieved snippets. We also observed a ``lost-in-the-middle'' phenomenon \citep{liu2023lost} between model performance and the position of the ground-truth snippet. 
Finally, we provide several practical recommendations based on the results and analyses, which can guide the application and future research of RAG in the biomedical domain.

In summary, our contributions are three-fold:
\begin{itemize}
    \item We introduce the \textsc{Mirage}\footnote{\url{https://github.com/Teddy-XiongGZ/MIRAGE}}, a first-of-its-kind benchmark for systematically comparing different medical RAG systems.
    \item We provide \textsc{MedRag}\footnote{\url{https://github.com/Teddy-XiongGZ/MedRAG}}, a RAG toolkit for medical QA that incorporates various domain-specific corpora, retrievers, and LLMs.
    \item We recommend a set of best practices for research and deployments of medical RAG systems based on our comprehensive results and analyses on \textsc{Mirage} with \textsc{MedRag}.
\end{itemize}

%% file: content/related_work.tex
\section{Related Work}
\subsection{Retrieval-augmented Generation}
Retrieval-Augmented Generation (RAG) was proposed by \citet{lewis2020retrieval} to enhance the generation performance on knowledge-intensive tasks by integrating retrieved relevant information. RAG not only mitigates the problem of hallucinations as LLMs are grounded on given contexts, but can also provide up-to-date knowledge that might not be encoded by the LLMs. Many follow-up studies have been carried out to improve over the vanilla RAG \cite{borgeaud2022improving,ram2023context,gao2023retrieval,jiang2023active, mialon2023augmented}.

In biomedicine, there have also been various explorations on how LLMs can improve literature information-seeking and clinical decision-making with RAG \cite{frisoni2022bioreader,naik2022literature,jin2023retrieve,lala2023paperqa,zakka2024almanac,jeong2024improving,wang2023augmenting}, but their evaluations are not comprehensive.
Nevertheless, current systematic evaluations in biomedicine typically focus on the vanilla LLMs without RAG \citep{chen2023large,nori2023capabilities}.
Our study provides the first systematic evaluations of RAG systems in medicine.

\subsection{Biomedical Question Answering}
Biomedical or medical question answering (QA) is a widely studied task since various information needs are expressed by natural language questions in biomedicine \cite{zweigenbaum2003question,athenikos2010biomedical,jin2022biomedical}.
While BERT-based \cite{devlin2019bert} models used to be the state-of-the-art methods of medical QA \cite{abacha2019overview,lee2020biobert,soni2020evaluation,gu2021domain,yasunaga2022deep}, they are outperformed by LLMs with large margins \cite{singhal2023towards,chen2023meditron,nori2023can}.
Due to their knowledge-intensive nature, QA datasets are commonly used to evaluate the biomedical capabilities of both general LLMs \citep{nori2023capabilities,nori2023can} and domain-specific LLMs  \cite{luo2022biogpt,chen2023meditron,wu2023pmc,singhal2023large,singhal2023towards}.
Following these studies, we also use medical QA datasets to test if a RAG system can retrieve and leverage relevant contexts. 
Unlike prior efforts, our evaluation employs both RAG and question-only retrieval settings, a more realistic evaluation for medical QA. 

%% file: content/benchmark.tex
\section{The \textsc{Mirage} Benchmark}

\subsection{Evaluation Settings}
The main objective of this work is to evaluate RAG systems in a setting that reflects real-world medical information needs as much as possible while being practically scalable.
As such, our \textsc{Mirage} benchmark adopts four key evaluation settings:

\paragraph{Zero-Shot Learning (ZSL).} As real-world medical questions are often posed without similar exemplars available, in our benchmark, the RAG systems should be evaluated in a zero-shot setting where in-context few-shot learning is not permitted.

\paragraph{Multi-Choice Evaluation (MCE).} Evaluating medical QA systems using multi-choice questions is a widely adopted method that can be practically implemented for large-scale evaluation \cite{nori2023capabilities,nori2023can,singhal2023large,lievin2022can,lala2023paperqa}. To be consistent with existing research, we also use a multi-choice setting in our benchmark to compare different systems.

\paragraph{Retrieval-Augmented Generation (RAG).} The medical questions used in \textsc{Mirage} are knowledge-intensive, which are difficult to answer without external knowledge. 
Moreover, due to the problem of hallucination, letting LLMs be reasoning engines instead of knowledge databases could be a better practice in medicine \cite{truhn2023large}.
For the above reasons, RAG should be utilized to collect external information for accurate and reliable answer generation. 

\paragraph{Question-Only Retrieval (QOR).} To align with real-world cases of medical QA, answer options should not be provided as input during retrieval. This is a more realistic setting for evaluating RAG systems. 
While \citet{lievin2022can} and \citet{lala2023paperqa} evaluated LLMs with RAG on medical QA, options were used for retrieval in their work, which is not a realistic setting. 
To the best of our knowledge, we are the first to propose and employ this setting for medical QA evaluation.

Table \ref{tab:setting_comp} lists related work on the evaluation settings. Only \textsc{Mirage} adopts all four considerations.

\begin{table}[h]\small
    \centering
    \begin{tabular}{lccccc}
    \toprule
        \textbf{Study} & \textbf{ZSL} & \textbf{MCE} & \textbf{RAG} & \textbf{QOR} \\
    \midrule
        \citet{nori2023capabilities} & \checkmark & \checkmark &  & \\
        \citet{singhal2023large} & \checkmark & \checkmark &  & \\
        \citet{lievin2022can} & \checkmark & \checkmark & \checkmark & \\
        \citet{lala2023paperqa} & \checkmark & \checkmark & \checkmark & \\
        \textsc{Mirage} (Ours) & \checkmark & \checkmark & \checkmark & \checkmark \\
    \bottomrule
    \end{tabular}
    \caption{Comparison of related work for using the different evaluation settings adopted in \textsc{Mirage}.}
    \label{tab:setting_comp}
\end{table}

\subsection{Component Datasets} \label{sec:bench_dataset}

\begin{figure}[ht]
    \centering
    \includegraphics[width=0.95\linewidth]{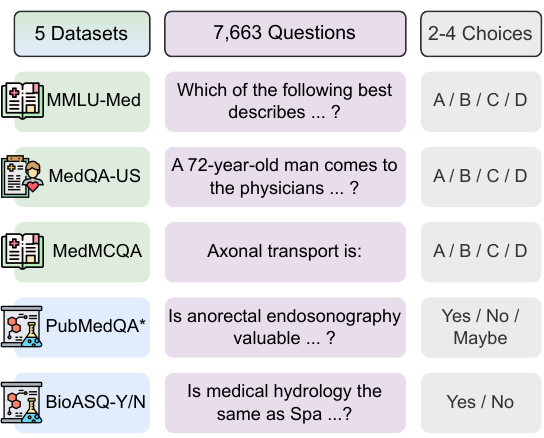}
    \caption{Composition of the \textsc{Mirage} benchmark.}
    \label{fig:MIRAGE}
\end{figure}

As shown in Figure \ref{fig:MIRAGE}, \textsc{Mirage} contains five commonly used datasets for medical QA for the evaluation of RAG systems \cite{hendrycks2020measuring,jin2021disease,pal2022medmcqa,jin2019pubmedqa,tsatsaronis2015overview}, including three medical examination QA datasets (MMLU-Med, MedQA-US, MedMCQA) and two biomedical research QA datasets (PubMedQA*, BioASQ-Y/N). Specifically, we only include multi-choice questions that are related to biomedicine and exclude all ground-truth supporting contexts for the questions.
For example, we remove the contexts of PubMedQA and only use the questions, resulting in PubMedQA*.
More details are described in the appendix. Table \ref{tab:task_statistic} presents the statistics of the datasets in \textsc{Mirage}.

As the tasks in \textsc{Mirage} are all composed of multi-choice questions, we evaluate a given RAG system by testing its performance in predicting the correct answer choices.
For each specific task, we compute the accuracy of model predictions as the evaluation metric, as well as the standard deviation for the proportion of correctly answered questions, reflecting the error bound of the results. Across all five tasks in \textsc{Mirage}, an average score of the accuracies will be measured to show how a given system performs on medical QA in general.

\begin{table}[h] \small
    \centering
    \begin{tabular}{lccccccccc}
        \toprule
        \bf Dataset & \bf Size & \bf \#O.  & \bf Avg. L & \bf Source \\
        \midrule
        MMLU-Med & 1,089 & 4 & 63 & Examination \\
        MedQA-US & 1,273 & 4 & 177 & Examination \\
        MedMCQA & 4,183 & 4 & 26 & Examination \\
        PubMedQA* & 500 & 3 & 24 & Literature \\
        BioASQ-Y/N & 618 & 2 & 17 & Literature \\
        \bottomrule
    \end{tabular}
    \caption{Statistics of \textsc{Mirage} tasks. \#O.: numbers of options; Avg. L: average token counts in each question.}
    \label{tab:task_statistic}
\end{table}

%% file: content/method.tex
\section{The \textsc{MedRag} Toolkit}
\begin{figure}[ht]
    \centering
    \includegraphics[width=1\linewidth]{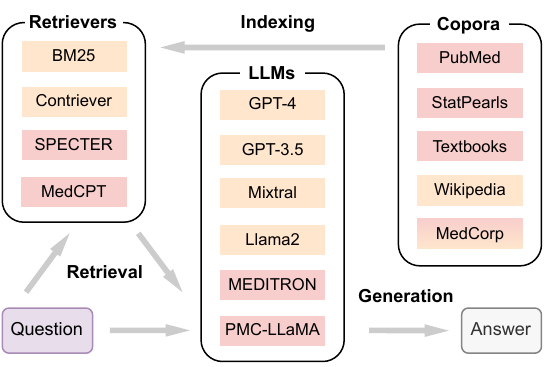}
    \caption{Component overview of the \textsc{MedRag} toolkit.}
    \label{fig:MedRAG}
\end{figure}

To comprehensively evaluate how different RAG systems perform on our \textsc{Mirage} benchmark, we propose \textsc{MedRag}, a toolkit with systematic implementations of RAG for medical QA. As shown in Figure \ref{fig:MedRAG}, \textsc{MedRag} consists of three major components: Corpora, Retrievers, and LLMs, which are briefly introduced in this section. More details of each component can be found in the appendix.

For corpora used in \textsc{MedRag}, we collect raw data from four different sources, including the commonly used PubMed\footnote{\url{https://pubmed.ncbi.nlm.nih.gov/}} for all biomedical abstracts, StatPearls\footnote{\url{https://www.statpearls.com/}} for clinical decision support, medical Textbooks \cite{jin2021disease} for domain-specific knowledge, and Wikipedia for general knowledge.
To the best of our knowledge, this is the first work that evaluates new corpora like StatPearls.
We also provide a MedCorp corpus by combining all four corpora, facilitating cross-source retrieval. Each corpus is chunked into short snippets. Statistics of used corpora are shown in Table \ref{tab:corpus_statistic}.

\begin{table}[ht] \small
    \centering
    \begin{tabular}{lccccccccc}
        \toprule
        \textbf{Corpus} & \bf \#Doc. & \textbf{\#Snippets} & \textbf{Avg. L} & \textbf{Domain} \\
        \midrule
        PubMed & 23.9M & 23.9M & 296 & Biomed. \\
        StatPearls & 9.3k  & 301.2k & 119 & Clinics \\
        Textbooks & 18 & 125.8k & 182 & Medicine \\
        Wikipedia & 6.5M & 29.9M & 162 & General \\
        MedCorp & 30.4M & 54.2M & 221 & Mixed \\
        \bottomrule
    \end{tabular}
    \caption{Statistics of corpora in \textsc{MedRag}. \#Doc.: numbers of raw documents; \#Snippets: numbers of snippets (chunks); Avg. L: average length of snippets.}
    \label{tab:corpus_statistic}
\end{table}

For the retrieval algorithms, while many general and domain-specific retrievers have been proposed \cite{remy2022biolord,ostendorff2022neighborhood,karpukhin2020dense,xiong2020approximate}, we only select some representative ones in \textsc{MedRag} due to limited resources, including a lexical retriever (BM25, \citeauthor{robertson2009probabilistic}, \citeyear{robertson2009probabilistic}), a general-domain semantic retriever (Contriever, \citeauthor{izacard2022unsupervised}, \citeyear{izacard2022unsupervised}), a scientific-domain retriever (SPECTER, \citeauthor{cohan2020specter}, \citeyear{cohan2020specter}), and a biomedical-domain retriever (MedCPT, \citeauthor{jin2023medcpt}, \citeyear{jin2023medcpt}). Their statistics are presented in Table \ref{tab:retriever_statistic}. In our experiments, 32 snippets are retrieved by default. Additionally, we utilize Reciprocal Rank Fusion (RRF, \citeauthor{cormack2009reciprocal}, \citeyear{cormack2009reciprocal}) to combine results from different retrievers, including RRF-2 (fusion of BM25 and MedCPT), and RRF-4 (fusion of all four retrievers).

\begin{table}[ht] \small
    \centering
    \begin{tabular}{lccccccccc}
        \toprule
        \bf Retriever & \bf Type & \bf Size & \bf Metric & \bf Domain\\
        \midrule
        BM25 & Lexical & -- & BM25 & General \\
        Contriever & Semantic & 110M & IP & General \\
        SPECTER & Semantic & 110M & L2 & Scientific \\
        MedCPT & Semantic & 109M & IP & Biomed. \\
        \bottomrule
    \end{tabular}
    \caption{Statistics of Retrievers in \textsc{MedRag}, where IP stands for inner product and L2 stands for L2 norm.}
    \label{tab:retriever_statistic}
\end{table}

Similarly, although various LLMs have emerged in recent years \cite{singhal2023large,singhal2023towards,taylor2022galactica,luo2022biogpt,yang2022gatortron}, we select several frequently used ones in \textsc{MedRag}, including the commercial GPT-3.5 and GPT-4 \cite{openai2023gpt4}, the open-source Mixtral \cite{jiang2024mixtral} and Llama2 \cite{touvron2023llama2}, and the biomedical domain-specific MEDITRON \cite{chen2023meditron} and PMC-LLaMA \cite{wu2023pmc}. Statistics of the used LLMs can be found in Table \ref{tab:llm_statistic}. For all LLMs, we concatenate and prepend retrieved snippets to the question input, and perform chain-of-thought (CoT) prompting \cite{wei2022chain} in \textsc{MedRag} to fully leverage the reasoning capability of the models. 
Temperatures are set to 0 for deterministic outputs. CoT without RAG is used as the baseline for comparison.

\begin{table}[h] \small
    \centering
    \begin{tabular}{lccccccccc}
        \toprule
        \bf LLM & \bf Size & \bf Context & \bf Open & \bf Domain\\
        \midrule
        GPT-4 & N/A & 32,768 & No & General \\
        GPT-3.5 & N/A & 16,384 & No & General \\
        Mixtral & 8\(\times\)7B & 32,768 & Yes & General \\
        Llama2 & 70B & 4,096 & Yes & General \\
        MEDITRON & 70B & 4,096 & Yes & Biomed. \\
        PMC-LLaMA & 13B & 2,048 & Yes & Biomed. \\
        \bottomrule
    \end{tabular}
    \caption{Statistics of LLMs used in \textsc{MedRag}. Context: context length of the LLM; Open: Open-source.}
    \label{tab:llm_statistic}
\end{table}

%% file: content/results.tex
\section{Results}

\begin{table*}[h]\small
\centering
\begin{tabular}{cccccccccc}
\toprule
\multirow{3}{*}{\bf LLM} & \multirow{3}{*}{\bf Method} & \multicolumn{5}{c}{\bf \textsc{Mirage} Benchmark Dataset} & \multirow{3}{*}{\bf Avg.} \\ 
\cmidrule(lr){3-7}
 &  & \bf MMLU-Med & \bf MedQA-US & \bf MedMCQA & \bf PubMedQA* & \bf BioASQ-Y/N &  \\
\midrule
\multirow{2}{*}{\makecell{\textbf{GPT-4} \\ (-32k-0613)}} 
& CoT & 89.44 \textcolor{gray}{\scriptsize $\pm$ 0.93} & 83.97 \textcolor{gray}{\scriptsize $\pm$ 1.03} & 69.88 \textcolor{gray}{\scriptsize $\pm$ 0.71} & 39.60 \textcolor{gray}{\scriptsize $\pm$ 2.19} & 84.30 \textcolor{gray}{\scriptsize $\pm$ 1.46} & 73.44 \\
& \textsc{MedRag} & 87.24 \textcolor{gray}{\scriptsize $\pm$ 1.01} & 82.80 \textcolor{gray}{\scriptsize $\pm$ 1.06} & 66.65 \textcolor{gray}{\scriptsize $\pm$ 0.73} & 70.60 \textcolor{gray}{\scriptsize $\pm$ 2.04} & 92.56 \textcolor{gray}{\scriptsize $\pm$ 1.06} & 79.97 \\
\midrule
\multirow{2}{*}{\makecell{\textbf{GPT-3.5}\\(-16k-0613)}} 
& CoT & 72.91 \textcolor{gray}{\scriptsize $\pm$ 1.35} & 65.04 \textcolor{gray}{\scriptsize $\pm$ 1.34} & 55.25 \textcolor{gray}{\scriptsize $\pm$ 0.77} & 36.00 \textcolor{gray}{\scriptsize $\pm$ 2.15} & 74.27 \textcolor{gray}{\scriptsize $\pm$ 1.76} & 60.69 \\
& \textsc{MedRag} & 75.48 \textcolor{gray}{\scriptsize $\pm$ 1.30} & 66.61 \textcolor{gray}{\scriptsize $\pm$ 1.32} & 58.04 \textcolor{gray}{\scriptsize $\pm$ 0.76} &  67.40 \textcolor{gray}{\scriptsize $\pm$ 2.10} & 90.29 \textcolor{gray}{\scriptsize $\pm$ 1.19} & 71.57 \\
\midrule
\multirow{2}{*}{ \makecell{\textbf{Mixtral}\\(8\(\times\)7B)} }
& CoT & 74.01 \textcolor{gray}{\scriptsize $\pm$ 1.33} & 64.10 \textcolor{gray}{\scriptsize $\pm$ 1.34} & 56.28 \textcolor{gray}{\scriptsize $\pm$ 0.77} & 35.20 \textcolor{gray}{\scriptsize $\pm$ 2.14} & 77.51 \textcolor{gray}{\scriptsize $\pm$ 1.68} & 61.42 \\
& \textsc{MedRag} & 75.85 \textcolor{gray}{\scriptsize $\pm$ 1.30} & 60.02 \textcolor{gray}{\scriptsize $\pm$ 1.37} & 56.42 \textcolor{gray}{\scriptsize $\pm$ 0.77} & 67.60 \textcolor{gray}{\scriptsize $\pm$ 2.09} & 87.54 \textcolor{gray}{\scriptsize $\pm$ 1.33} & 69.48 \\
\midrule
\multirow{2}{*}{ \makecell{\textbf{Llama2}\\(70B)} }
& CoT & 57.39 \textcolor{gray}{\scriptsize $\pm$ 1.50} & 47.84 \textcolor{gray}{\scriptsize $\pm$ 1.40} & 42.60 \textcolor{gray}{\scriptsize $\pm$ 0.76} & 42.20 \textcolor{gray}{\scriptsize $\pm$ 2.21} & 61.17 \textcolor{gray}{\scriptsize $\pm$ 1.96} & 50.24 \\
& \textsc{MedRag} & 54.55 \textcolor{gray}{\scriptsize $\pm$ 1.51} & 44.93 \textcolor{gray}{\scriptsize $\pm$ 1.39} & 43.08 \textcolor{gray}{\scriptsize $\pm$ 0.77} & 50.40 \textcolor{gray}{\scriptsize $\pm$ 2.24} & 73.95 \textcolor{gray}{\scriptsize $\pm$ 1.77} & 53.38 \\
\midrule
\multirow{2}{*}{\makecell{\textbf{MEDITRON}\\(70B)}} 
& CoT & 64.92 \textcolor{gray}{\scriptsize $\pm$ 1.45} & 51.69 \textcolor{gray}{\scriptsize $\pm$ 1.40} & 46.74 \textcolor{gray}{\scriptsize $\pm$ 0.77} & 53.40 \textcolor{gray}{\scriptsize $\pm$ 2.23} & 68.45 \textcolor{gray}{\scriptsize $\pm$ 1.87} & 57.04 \\
& \textsc{MedRag} & 65.38 \textcolor{gray}{\scriptsize $\pm$ 1.44} & 49.57 \textcolor{gray}{\scriptsize $\pm$ 1.40} & 52.67 \textcolor{gray}{\scriptsize $\pm$ 0.77} & 56.40 \textcolor{gray}{\scriptsize $\pm$ 2.22} & 76.86 \textcolor{gray}{\scriptsize $\pm$ 1.70} & 60.18 \\
\midrule
\multirow{2}{*}{\makecell{\textbf{PMC-LLaMA}\\(13B)}} 
& CoT & 52.16 \textcolor{gray}{\scriptsize $\pm$ 1.51} & 44.38 \textcolor{gray}{\scriptsize $\pm$ 1.39} & 46.55 \textcolor{gray}{\scriptsize $\pm$ 0.77} & 55.80 \textcolor{gray}{\scriptsize $\pm$ 2.22} & 63.11 \textcolor{gray}{\scriptsize $\pm$ 1.94} & 52.40 \\
& \textsc{MedRag} & 52.53 \textcolor{gray}{\scriptsize $\pm$ 1.51} & 42.58 \textcolor{gray}{\scriptsize $\pm$ 1.39} & 48.29 \textcolor{gray}{\scriptsize $\pm$ 0.77} & 56.00 \textcolor{gray}{\scriptsize $\pm$ 2.22} & 65.21 \textcolor{gray}{\scriptsize $\pm$ 1.92} & 52.92 \\
\bottomrule
\end{tabular}
\caption{Benchmark results of different backbone LLMs on \textsc{Mirage}. All numbers are accuracy in percentages.}
\label{tab:model_comp}
\end{table*}

We systematically evaluate \textsc{MedRag} on our \textsc{Mirage} benchmark, which provides us with a multi-dimensional analysis of different components in RAG for medicine. Section \ref{sec:res_llm} presents the results for different LLMs, and Section \ref{sec:res_corp_ret} includes the results of different corpora and retrievers.

\subsection{Comparison of Backbone LLMs} \label{sec:res_llm}
We first benchmark various LLMs on \textsc{Mirage} under both the CoT and the \textsc{MedRag} settings. 
For different LLMs, we use the same MedCorp corpus and the RRF-4 retriever and prepend 32 retrieved snippets for RAG. Results are shown in Table \ref{tab:model_comp}. 

Under the CoT setting, GPT-4 significantly outperforms other competitors, with an average score of 73.44\% on \textsc{Mirage}. While the best average score of other backbone LLMs can only achieve about 61\% (GPT-3.5 and Mixtral) in the CoT setting, their performance can be significantly improved to around 70\% with \textsc{MedRag}, which is comparable to GPT-4 (CoT). These results suggest the great potential of RAG as a way to enhance the zero-shot capability of LLMs to answer medical questions, which can be a more efficient choice than performing larger-scale pre-training. 
On all five tasks in \textsc{Mirage}, Mixtral shows an accuracy of 61.42\% on average in the CoT setting, which slightly surpasses the performance of GPT-3.5. However, Mixtral is still outperformed by GPT-3.5 with \textsc{MedRag} by 3.0\%, indicating the advantage of GPT-3.5 in following \textsc{MedRag} instructions.

Our results also demonstrate that domain-specific LLMs can exhibit advantages in certain cases. For example, in the CoT setting for PubMedQA*, MEDITRON and PMC-LLaMA present significantly higher accuracies than all other models, including GPT-4 (+34.8\% \& +40.9\%).
Additionally, MEDITRON shows a better performance in both CoT (+13.5\%) and \textsc{MedRag} (+12.7\%) than its base Llama2 model.  The comparison of Llama2 (\textsc{MedRag}) and MEDITRON (CoT) reflects the differences between RAG (+6.3\%) and supervised fine-tuning (SFT, +13.5\%) in improving the performance of LLMs on medical QA. While SFT is better at fusing medical knowledge into LLMs, RAG remains a more flexible and cost-efficient way to improve medical QA. For questions in PubMedQA* and BioASQ-Y/N where the closely related literature can be found from PubMed, \textsc{MedRag} greatly improves the ability of Llama2 to answer medical questions (+19.4\% \& +20.9\%), leading to a comparable or even better performance than MEDITRON (CoT). 
However, for examination questions in \textsc{Mirage} that are carefully designed to differentiate between medical students, \textsc{MedRag} does not always improve over SFT since the helpful snippets might be difficult to retrieve. The performance gap between these two types of questions suggests that there is still much room for improvement.

\subsection{Comparison of Corpora and Retrievers} \label{sec:res_corp_ret}

\begin{table*}[h]\small
\centering
\begin{tabular}{cccccccccc}
\toprule
\multirow{3}{*}{\bf Corpus} & \multirow{3}{*}{\bf Retriever} & \multicolumn{5}{c}{\bf \textsc{Mirage} Benchmark Dataset} & \multirow{3}{*}{\bf Average} \\ 
\cmidrule(lr){3-7}
 &  & \bf MMLU-Med & \bf MedQA-US & \bf MedMCQA & \bf PubMedQA* & \bf BioASQ-Y/N &  \\
\midrule
None & None & 72.91 \textcolor{gray}{\scriptsize $\pm$ 1.35} & 65.04 \textcolor{gray}{\scriptsize $\pm$ 1.34} & 55.25 \textcolor{gray}{\scriptsize $\pm$ 0.77} & 36.00 \textcolor{gray}{\scriptsize $\pm$ 2.15} & 74.27 \textcolor{gray}{\scriptsize $\pm$ 1.76} & 60.69 \\
\midrule
\multirow{6}{*}{\makecell{\textbf{PubMed}\\ (23.9M)}} 
& BM25 & \cellcolor{red!40.4!} 72.27 \textcolor{gray}{\scriptsize $\pm$ 1.36} & \cellcolor{red!67.5!} 63.71 \textcolor{gray}{\scriptsize $\pm$ 1.35} & \cellcolor{green!6.8!} 55.49 \textcolor{gray}{\scriptsize $\pm$ 0.77} & \cellcolor{green!73.2!} 66.20 \textcolor{gray}{\scriptsize $\pm$ 2.12} & \cellcolor{green!71.1!} 88.51 \textcolor{gray}{\scriptsize $\pm$ 1.28} & \cellcolor{green!62.8!} 69.23 \\
& Contriever & \cellcolor{red!74.1!} 71.72 \textcolor{gray}{\scriptsize $\pm$ 1.36} & \cellcolor{red!55.5!} 63.94 \textcolor{gray}{\scriptsize $\pm$ 1.35} & \cellcolor{red!30.1!} 54.29 \textcolor{gray}{\scriptsize $\pm$ 0.77} & \cellcolor{green!71.7!} 65.60 \textcolor{gray}{\scriptsize $\pm$ 2.12} & \cellcolor{green!55.7!} 85.44 \textcolor{gray}{\scriptsize $\pm$ 1.42} & \cellcolor{green!55.2!} 68.20 \\
& SPECTER & \cellcolor{green!5.5!} 73.19 \textcolor{gray}{\scriptsize $\pm$ 1.34} & \cellcolor{green!5.2!} 65.20 \textcolor{gray}{\scriptsize $\pm$ 1.34} & \cellcolor{red!66.8!} 53.12 \textcolor{gray}{\scriptsize $\pm$ 0.77} & \cellcolor{green!45.6!} 54.80 \textcolor{gray}{\scriptsize $\pm$ 2.23} & \cellcolor{green!7.3!} 75.73 \textcolor{gray}{\scriptsize $\pm$ 1.72} & \cellcolor{green!27.3!} 64.41 \\
& MedCPT & \cellcolor{green!3.5!} 73.09 \textcolor{gray}{\scriptsize $\pm$ 1.34} & \cellcolor{green!54.0!} 66.69 \textcolor{gray}{\scriptsize $\pm$ 1.32} & \cellcolor{red!9.8!} 54.94 \textcolor{gray}{\scriptsize $\pm$ 0.77} & \cellcolor{green!73.7!} 66.40 \textcolor{gray}{\scriptsize $\pm$ 2.11} & \cellcolor{green!57.3!} 85.76 \textcolor{gray}{\scriptsize $\pm$ 1.41} & \cellcolor{green!63.8!} 69.38 \\
& RRF-2 & \cellcolor{green!56.3!} 75.57 \textcolor{gray}{\scriptsize $\pm$ 1.30} & \cellcolor{red!35.6!} 64.34 \textcolor{gray}{\scriptsize $\pm$ 1.34} & \cellcolor{green!2.7!} 55.34 \textcolor{gray}{\scriptsize $\pm$ 0.77} & \cellcolor{green!80.0!} 69.00 \textcolor{gray}{\scriptsize $\pm$ 2.07} & \cellcolor{green!63.8!} 87.06 \textcolor{gray}{\scriptsize $\pm$ 1.35} & \cellcolor{green!70.3!} 70.26 \\
& RRF-4 & \cellcolor{green!9.4!} 73.37 \textcolor{gray}{\scriptsize $\pm$ 1.34} & \cellcolor{red!15.7!} 64.73 \textcolor{gray}{\scriptsize $\pm$ 1.34} & \cellcolor{red!15.8!} 54.75 \textcolor{gray}{\scriptsize $\pm$ 0.77} & \cellcolor{green!75.6!} 67.20 \textcolor{gray}{\scriptsize $\pm$ 2.10} & \cellcolor{green!71.1!} 88.51 \textcolor{gray}{\scriptsize $\pm$ 1.28} & \cellcolor{green!66.3!} 69.71 \\
\midrule
\multirow{6}{*}{\makecell{\textbf{StatPearls}\\(301.2k)}} 
& BM25 & \cellcolor{red!79.7!} 71.63 \textcolor{gray}{\scriptsize $\pm$ 1.37} & \cellcolor{green!20.6!} 65.67 \textcolor{gray}{\scriptsize $\pm$ 1.33} & \cellcolor{red!11.3!} 54.89 \textcolor{gray}{\scriptsize $\pm$ 0.77} & \cellcolor{red!48.7!} 27.60 \textcolor{gray}{\scriptsize $\pm$ 2.00} & \cellcolor{red!59.3!} 60.36 \textcolor{gray}{\scriptsize $\pm$ 1.97} & \cellcolor{red!59.7!} 56.03 \\
& Contriever & \cellcolor{green!7.4!} 73.28 \textcolor{gray}{\scriptsize $\pm$ 1.34} & \cellcolor{green!79.6!} 67.48 \textcolor{gray}{\scriptsize $\pm$ 1.31} & \cellcolor{red!31.6!} 54.24 \textcolor{gray}{\scriptsize $\pm$ 0.77} & \cellcolor{red!41.7!} 28.80 \textcolor{gray}{\scriptsize $\pm$ 2.03} & \cellcolor{red!67.5!} 58.41 \textcolor{gray}{\scriptsize $\pm$ 1.98} & \cellcolor{red!54.4!} 56.44 \\
& SPECTER & \cellcolor{green!17.2!} 73.74 \textcolor{gray}{\scriptsize $\pm$ 1.33} & \cellcolor{red!15.7!} 64.73 \textcolor{gray}{\scriptsize $\pm$ 1.34} & \cellcolor{red!75.8!} 52.83 \textcolor{gray}{\scriptsize $\pm$ 0.77} & \cellcolor{red!74.1!} 23.20 \textcolor{gray}{\scriptsize $\pm$ 1.89} & \cellcolor{red!70.3!} 57.77 \textcolor{gray}{\scriptsize $\pm$ 1.99} & \cellcolor{red!79.8!} 54.45 \\
& MedCPT & \cellcolor{red!6.8!} 72.82 \textcolor{gray}{\scriptsize $\pm$ 1.35} & \cellcolor{red!7.8!} 64.89 \textcolor{gray}{\scriptsize $\pm$ 1.34} & \cellcolor{red!33.8!} 54.17 \textcolor{gray}{\scriptsize $\pm$ 0.77} & \cellcolor{red!48.7!} 27.60 \textcolor{gray}{\scriptsize $\pm$ 2.00} & \cellcolor{red!57.9!} 60.68 \textcolor{gray}{\scriptsize $\pm$ 1.96} & \cellcolor{red!59.6!} 56.03 \\
& RRF-2 & \cellcolor{red!18.0!} 72.64 \textcolor{gray}{\scriptsize $\pm$ 1.35} & \cellcolor{green!20.6!} 65.67 \textcolor{gray}{\scriptsize $\pm$ 1.33} & \cellcolor{red!19.6!} 54.63 \textcolor{gray}{\scriptsize $\pm$ 0.77} & \cellcolor{red!34.8!} 30.00 \textcolor{gray}{\scriptsize $\pm$ 2.05} & \cellcolor{red!55.8!} 61.17 \textcolor{gray}{\scriptsize $\pm$ 1.96} & \cellcolor{red!49.5!} 56.82 \\
& RRF-4 & \cellcolor{green!19.1!} 73.83 \textcolor{gray}{\scriptsize $\pm$ 1.33} & \cellcolor{green!2.7!} 65.12 \textcolor{gray}{\scriptsize $\pm$ 1.34} & \cellcolor{red!45.1!} 53.81 \textcolor{gray}{\scriptsize $\pm$ 0.77} & \cellcolor{red!31.3!} 30.60 \textcolor{gray}{\scriptsize $\pm$ 2.06} & \cellcolor{red!62.0!} 59.71 \textcolor{gray}{\scriptsize $\pm$ 1.97} & \cellcolor{red!52.2!} 56.61 \\
\midrule
\multirow{6}{*}{\makecell{\textbf{Textbooks}\\(125.8k)}} 
& BM25 & \cellcolor{green!36.7!} 74.66 \textcolor{gray}{\scriptsize $\pm$ 1.32} & \cellcolor{green!48.8!} 66.54 \textcolor{gray}{\scriptsize $\pm$ 1.32} & \cellcolor{red!37.6!} 54.05 \textcolor{gray}{\scriptsize $\pm$ 0.77} & \cellcolor{red!33.6!} 30.20 \textcolor{gray}{\scriptsize $\pm$ 2.05} & \cellcolor{red!60.7!} 60.03 \textcolor{gray}{\scriptsize $\pm$ 1.97} & \cellcolor{red!46.0!} 57.10 \\
& Contriever & \cellcolor{green!25.0!} 74.10 \textcolor{gray}{\scriptsize $\pm$ 1.33} & \cellcolor{green!69.4!} 67.16 \textcolor{gray}{\scriptsize $\pm$ 1.32} & \cellcolor{red!22.6!} 54.53 \textcolor{gray}{\scriptsize $\pm$ 0.77} & \cellcolor{red!54.5!} 26.60 \textcolor{gray}{\scriptsize $\pm$ 1.98} & \cellcolor{red!60.0!} 60.19 \textcolor{gray}{\scriptsize $\pm$ 1.97} & \cellcolor{red!53.4!} 56.52 \\
& SPECTER & \cellcolor{red!6.8!} 72.82 \textcolor{gray}{\scriptsize $\pm$ 1.35} & \cellcolor{green!77.1!} 67.40 \textcolor{gray}{\scriptsize $\pm$ 1.31} & \cellcolor{red!61.6!} 53.29 \textcolor{gray}{\scriptsize $\pm$ 0.77} & \cellcolor{red!60.2!} 25.60 \textcolor{gray}{\scriptsize $\pm$ 1.95} & \cellcolor{red!80.0!} 55.50 \textcolor{gray}{\scriptsize $\pm$ 2.00} & \cellcolor{red!73.8!} 54.92 \\
& MedCPT & \cellcolor{green!42.6!} 74.93 \textcolor{gray}{\scriptsize $\pm$ 1.31} & \cellcolor{green!38.6!} 66.22 \textcolor{gray}{\scriptsize $\pm$ 1.33} & \cellcolor{red!26.3!} 54.41 \textcolor{gray}{\scriptsize $\pm$ 0.77} & \cellcolor{red!39.4!} 29.20 \textcolor{gray}{\scriptsize $\pm$ 2.03} & \cellcolor{red!55.1!} 61.33 \textcolor{gray}{\scriptsize $\pm$ 1.96} & \cellcolor{red!44.4!} 57.22 \\
& RRF-2 & \cellcolor{green!79.7!} 76.68 \textcolor{gray}{\scriptsize $\pm$ 1.28} & \cellcolor{green!28.3!} 65.91 \textcolor{gray}{\scriptsize $\pm$ 1.33} & \cellcolor{red!14.3!} 54.79 \textcolor{gray}{\scriptsize $\pm$ 0.77} & \cellcolor{red!29.0!} 31.00 \textcolor{gray}{\scriptsize $\pm$ 2.07} & \cellcolor{red!63.4!} 59.39 \textcolor{gray}{\scriptsize $\pm$ 1.98} & \cellcolor{red!40.2!} 57.55 \\
& RRF-4 & \cellcolor{green!60.2!} 75.76 \textcolor{gray}{\scriptsize $\pm$ 1.30} & \cellcolor{green!33.5!} 66.06 \textcolor{gray}{\scriptsize $\pm$ 1.33} & \cellcolor{green!8.8!} 55.56 \textcolor{gray}{\scriptsize $\pm$ 0.77} & \cellcolor{red!32.4!} 30.40 \textcolor{gray}{\scriptsize $\pm$ 2.06} & \cellcolor{red!57.9!} 60.68 \textcolor{gray}{\scriptsize $\pm$ 1.96} & \cellcolor{red!38.4!} 57.69 \\
\midrule
\multirow{6}{*}{\makecell{\textbf{Wikipedia}\\(29.9M)}} 
& BM25 & \cellcolor{green!9.4!} 73.37 \textcolor{gray}{\scriptsize $\pm$ 1.34} & \cellcolor{red!79.4!} 63.47 \textcolor{gray}{\scriptsize $\pm$ 1.35} & \cellcolor{red!36.1!} 54.10 \textcolor{gray}{\scriptsize $\pm$ 0.77} & \cellcolor{red!55.6!} 26.40 \textcolor{gray}{\scriptsize $\pm$ 1.97} & \cellcolor{red!12.4!} 71.36 \textcolor{gray}{\scriptsize $\pm$ 1.82} & \cellcolor{red!37.8!} 57.74 \\
& Contriever & \cellcolor{green!25.0!} 74.10 \textcolor{gray}{\scriptsize $\pm$ 1.33} & \cellcolor{green!30.9!} 65.99 \textcolor{gray}{\scriptsize $\pm$ 1.33} & \cellcolor{red!38.3!} 54.03 \textcolor{gray}{\scriptsize $\pm$ 0.77} & \cellcolor{red!55.6!} 26.40 \textcolor{gray}{\scriptsize $\pm$ 1.97} & \cellcolor{red!18.6!} 69.90 \textcolor{gray}{\scriptsize $\pm$ 1.85} & \cellcolor{red!33.4!} 58.08 \\
& SPECTER & \cellcolor{red!46.0!} 72.18 \textcolor{gray}{\scriptsize $\pm$ 1.36} & \cellcolor{red!71.4!} 63.63 \textcolor{gray}{\scriptsize $\pm$ 1.35} & \cellcolor{red!79.6!} 52.71 \textcolor{gray}{\scriptsize $\pm$ 0.77} & \cellcolor{red!79.9!} 22.20 \textcolor{gray}{\scriptsize $\pm$ 1.86} & \cellcolor{red!31.7!} 66.83 \textcolor{gray}{\scriptsize $\pm$ 1.89} & \cellcolor{red!66.3!} 55.51 \\
& MedCPT & \cellcolor{red!57.2!} 71.99 \textcolor{gray}{\scriptsize $\pm$ 1.36} & \cellcolor{green!2.7!} 65.12 \textcolor{gray}{\scriptsize $\pm$ 1.34} & \cellcolor{red!3.1!} 55.15 \textcolor{gray}{\scriptsize $\pm$ 0.77} & \cellcolor{red!40.6!} 29.00 \textcolor{gray}{\scriptsize $\pm$ 2.03} & \cellcolor{red!3.4!} 73.46 \textcolor{gray}{\scriptsize $\pm$ 1.78} & \cellcolor{red!22.3!} 58.95 \\
& RRF-2 & \cellcolor{green!26.9!} 74.20 \textcolor{gray}{\scriptsize $\pm$ 1.33} & \cellcolor{red!23.7!} 64.57 \textcolor{gray}{\scriptsize $\pm$ 1.34} & \cellcolor{red!16.6!} 54.72 \textcolor{gray}{\scriptsize $\pm$ 0.77} & \cellcolor{red!29.0!} 31.00 \textcolor{gray}{\scriptsize $\pm$ 2.07} & \cellcolor{green!9.7!} 76.21 \textcolor{gray}{\scriptsize $\pm$ 1.71} & \cellcolor{red!7.0!} 60.14 \\
& RRF-4 & \cellcolor{green!5.5!} 73.19 \textcolor{gray}{\scriptsize $\pm$ 1.34} & \cellcolor{red!3.8!} 64.96 \textcolor{gray}{\scriptsize $\pm$ 1.34} & \cellcolor{red!22.6!} 54.53 \textcolor{gray}{\scriptsize $\pm$ 0.77} & \cellcolor{red!29.0!} 31.00 \textcolor{gray}{\scriptsize $\pm$ 2.07} & \cellcolor{red!9.6!} 72.01 \textcolor{gray}{\scriptsize $\pm$ 1.81} & \cellcolor{red!19.9!} 59.14 \\
\midrule
\multirow{6}{*}{\makecell{\textbf{MedCorp}\\(65.3M)}} 
& BM25 & \cellcolor{green!15.2!} 73.65 \textcolor{gray}{\scriptsize $\pm$ 1.34} & \cellcolor{green!28.3!} 65.91 \textcolor{gray}{\scriptsize $\pm$ 1.33} & \cellcolor{green!43.6!} 56.78 \textcolor{gray}{\scriptsize $\pm$ 0.77} & \cellcolor{green!73.2!} 66.20 \textcolor{gray}{\scriptsize $\pm$ 2.12} & \cellcolor{green!67.0!} 87.70 \textcolor{gray}{\scriptsize $\pm$ 1.32} & \cellcolor{green!68.7!} 70.05 \\
& Contriever & \cellcolor{green!54.3!} 75.48 \textcolor{gray}{\scriptsize $\pm$ 1.30} & \cellcolor{red!47.6!} 64.10 \textcolor{gray}{\scriptsize $\pm$ 1.34} & \cellcolor{green!24.5!} 56.11 \textcolor{gray}{\scriptsize $\pm$ 0.77} & \cellcolor{green!64.0!} 62.40 \textcolor{gray}{\scriptsize $\pm$ 2.17} & \cellcolor{green!53.3!} 84.95 \textcolor{gray}{\scriptsize $\pm$ 1.44} & \cellcolor{green!58.2!} 68.61 \\
& SPECTER & \cellcolor{green!30.9!} 74.38 \textcolor{gray}{\scriptsize $\pm$ 1.32} & \cellcolor{green!12.9!} 65.44 \textcolor{gray}{\scriptsize $\pm$ 1.33} & \cellcolor{red!26.3!} 54.41 \textcolor{gray}{\scriptsize $\pm$ 0.77} & \cellcolor{green!48.0!} 55.80 \textcolor{gray}{\scriptsize $\pm$ 2.22} & \cellcolor{red!4.8!} 73.14 \textcolor{gray}{\scriptsize $\pm$ 1.78} & \cellcolor{green!29.0!} 64.63 \\
& MedCPT & \cellcolor{green!38.7!} 74.75 \textcolor{gray}{\scriptsize $\pm$ 1.32} & \cellcolor{green!77.1!} 67.40 \textcolor{gray}{\scriptsize $\pm$ 1.31} & \cellcolor{green!17.0!} 55.85 \textcolor{gray}{\scriptsize $\pm$ 0.77} & \cellcolor{green!73.7!} 66.40 \textcolor{gray}{\scriptsize $\pm$ 2.11} & \cellcolor{green!58.2!} 85.92 \textcolor{gray}{\scriptsize $\pm$ 1.40} & \cellcolor{green!68.9!} 70.06 \\
& RRF-2 & \cellcolor{green!17.2!} 73.74 \textcolor{gray}{\scriptsize $\pm$ 1.33} & \cellcolor{green!71.9!} 67.24 \textcolor{gray}{\scriptsize $\pm$ 1.32} & \cellcolor{green!23.8!} 56.08 \textcolor{gray}{\scriptsize $\pm$ 0.77} & \cellcolor{green!77.1!} 67.80 \textcolor{gray}{\scriptsize $\pm$ 2.09} & \cellcolor{green!69.5!} 88.19 \textcolor{gray}{\scriptsize $\pm$ 1.30} & \cellcolor{green!72.9!} 70.61 \\
& RRF-4 & \cellcolor{green!54.3!} 75.48 \textcolor{gray}{\scriptsize $\pm$ 1.30} & \cellcolor{green!51.4!} 66.61 \textcolor{gray}{\scriptsize $\pm$ 1.32} & \cellcolor{green!79.8!} 58.04 \textcolor{gray}{\scriptsize $\pm$ 0.76} & \cellcolor{green!76.1!} 67.40 \textcolor{gray}{\scriptsize $\pm$ 2.10} & \cellcolor{green!80.0!} 90.29 \textcolor{gray}{\scriptsize $\pm$ 1.19} & \cellcolor{green!79.9!} 71.57 \\
\bottomrule
\end{tabular}
\caption{Accuracy (\%) of GPT-3.5 (\textsc{MedRag}) with different corpora and retrievers on \textsc{Mirage}. Red and green denote performance \colorbox{red!50}{decreases} and \colorbox{green!50}{increases} compared to CoT (first row). The shade reflects the relative change.}
\label{tab:retrieval_comp}
\end{table*}

We also compare how different corpora and retrievers affect the \textsc{Mirage} performance with \textsc{MedRag}. Based on the results in Table \ref{tab:model_comp}, we conduct the following experiments with GPT-3.5 as it benefits the most from \textsc{MedRag} (+17.9\%). 

As shown in Table \ref{tab:retrieval_comp}, the performance of one RAG system is strongly related to the corpus it selects. 
\textsc{MedRag} with Textbooks achieves the highest accuracy on MMLU-Med (76.68\%) and the one with StatPearls performs the best on MedQA-US (67.48\%).
However, these two corpora provide little assistance in answering questions from PubMedQA* and BioASQ-Y/N, which almost solely benefit from the PubMed corpus. This is expected due to the design of these two datasets. Overall, PubMed is the only corpus that provides improvement for all \textsc{Mirage} tasks, probably due to its large scale and domain-specificity. Therefore, selecting a suitable corpus for the task should be the first key step in RAG for medicine. While choosing task-specific corpora may require expert knowledge, we find MedCorp, a simple combination of all corpora, that performs robustly across various tasks, to be a satisfactory solution. As for the four tasks mentioned above, \textsc{MedRag} can always find useful snippets from the MedCorp corpus. Even on MedMCQA, where \textsc{MedRag} does not benefit from any single corpus, MedCorp still improves almost all retrievers (-1.5\% \(\sim\) +5.0\%).

\begin{figure*}[h]
    \centering
    \includegraphics[width=1\linewidth]{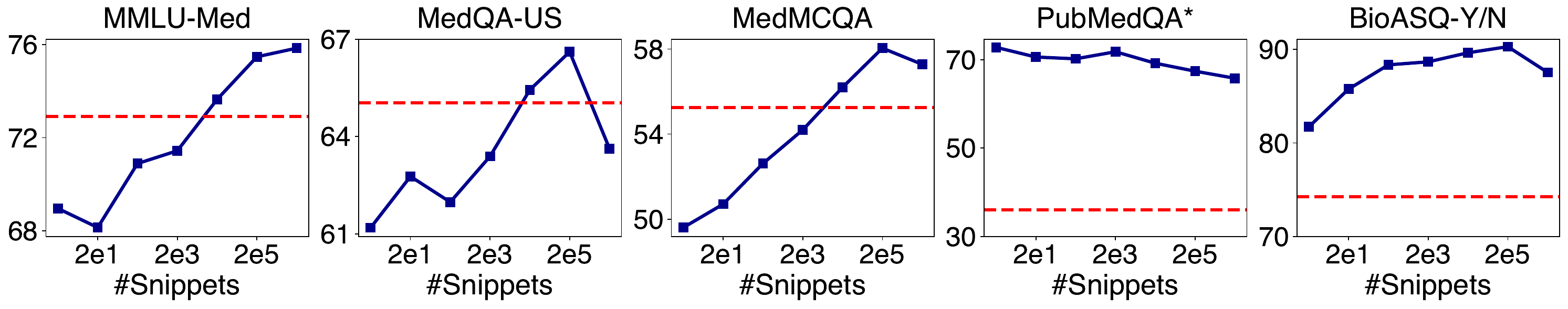}
    \caption{\textsc{MedRag} accuracy with different numbers of retrieved snippets. Red dotted lines denote CoT performance.}
    \label{fig:scaling}
\end{figure*}

The selection of retrievers is another flexibility in \textsc{MedRag} that affects overall performance, which decides whether relevant information can be found from corpora. Table \ref{tab:retrieval_comp} shows the variable performance of different retrievers, which can be explained by the data and strategy differences in their training. 
For example, MedCPT is a biomedical retriever that has been trained on PubMed user logs. Thus, compared with other retrievers, it has a better performance when PubMed is used as the corpus in \textsc{MedRag} (+0.2\% \(\sim\) +7.7\%). 
Similarly, with Wikipedia as part of the training data, Contriever shows better performance than other retrievers in tasks with the Wikipedia corpus, especially on MMLU-Med and MedQA-US. Moreover, during the training of SPECTER, the retriever is tuned to regularize pairwise article distances rather than query-to-article distances. As such, it has an inferior average performance to other individual retrievers (-7.8\% \(\sim\) -6.8\%) on MedCorp as its training setting mismatches the cases in medical QA.

Table \ref{tab:retrieval_comp} also shows that the fusion of retrieval results with RRF effectively improves the performance on \textsc{Mirage}. Using MedCorp, \textsc{MedRag} with RRF-4 have a 1.4\% to 10.7\% increase in the average performance compared to individual retrievers.
However, the fusion of more retrievers may not always lead to a better performance. For example, on Wikipedia where SPECTER has a poor performance across all tasks, RRF-2 shows a better average performance than RRF-4 on \textsc{Mirage} (+1.7\%). Specifically, for tasks like BioASQ-Y/N where both Contriever and SPECTER perform poorly, RRF-2 can significantly improve the performance of \textsc{MedRag}, which is better than RRF-4 (+5.8\%) and all other individual retrievers (+3.7\% \(\sim\) +14.0\%). In contrast, on MedQA-US where Contriever achieves the best score (65.99\%), RRF-2 underperforms RRF-4 (-0.6\%). 
On the MedCorp corpus where \textsc{MedRag} can benefit from all retrievers, RRF-4 brings a larger improvement than RRF-2, with a state-of-the-art average score of 71.57\% on our \textsc{Mirage} benchmark.

%% file: content/discussion.tex
\section{Discussions}

\subsection{Performance Scaling} \label{sec:scaling}
We explore how the performance of \textsc{MedRag} scales with the increase in the number of snippets used for medical QA. To study the scaling properties, we use GPT-3.5 as the backbone LLM, RRF-4 as the retriever, and MedCorp as the corpus. 

Figure \ref{fig:scaling} shows the scaling curves of \textsc{MedRag} on each task in \textsc{Mirage} with different numbers of snippets \(k\in\{1,2,4,...,64\}\). On MMLU-Med, MedQA-US, and MedMCQA, we see roughly log-linear curves in the scaling plots for \(k \leq 32\). The results show that when \(k\) is small (\(k\leq 8\) in this case), \textsc{MedRag} cannot provide enough useful information, which even hinders the LLM from using its inherent knowledge to derive the correct answer. 
In general, the RAG performance improves as \(k\) increases, indicating the existence of helpful knowledge from the retrieved snippets. However, the RAG performance can drop when \(k\) is too large and the signal-noise-ratio begins to decrease.

Compared with the three examination tasks, PubMedQA* and BioASQ-Y/N can be relatively easier for \textsc{MedRag} since the ground-truth supporting information can be found in PubMed. Figure \ref{fig:scaling} reveals that \textsc{MedRag} can achieve high accuracy on PubMedQA* with just \(k=1\), and its performance drops with the increase of \(k\) as more irrelevant snippets are entered, which corresponds to the fact that 79.6\% ground-truth snippets are successfully identified as the top-1 related context by the retrieval system. \textsc{MedRag} also shows a dramatic increase in accuracy on BioASQ-Y/N when \(k=1\), whose performance continues to grow as \(k\) gets larger. 

\subsection{Position of Ground-truth Snippet}

\citet{liu2023lost} found the RAG performance is lowest when the relevant information is placed in the middle, a phenomenon known as ``lost-in-the-middle''.
In our \textsc{Mirage} benchmark, PubMedQA* and BioASQ-Y/N are the tasks that have ground-truth labels of the supporting snippets for each question. Here we use PubMed as the corpus, and take GPT-3.5 and RRF-4 as the LLM and retriever, respectively. For each dataset, we group the positions of ground-truth snippets into several bins, on which we evaluate how accurate \textsc{MedRag} is in answering questions whose ground-truth snippets are in corresponding bins. For PubMedQA*, we only show the results of the first 18 positions, since no ground-truth snippets have been placed after it.

\begin{figure}[t!]
    \centering
    \includegraphics[width=1\linewidth]{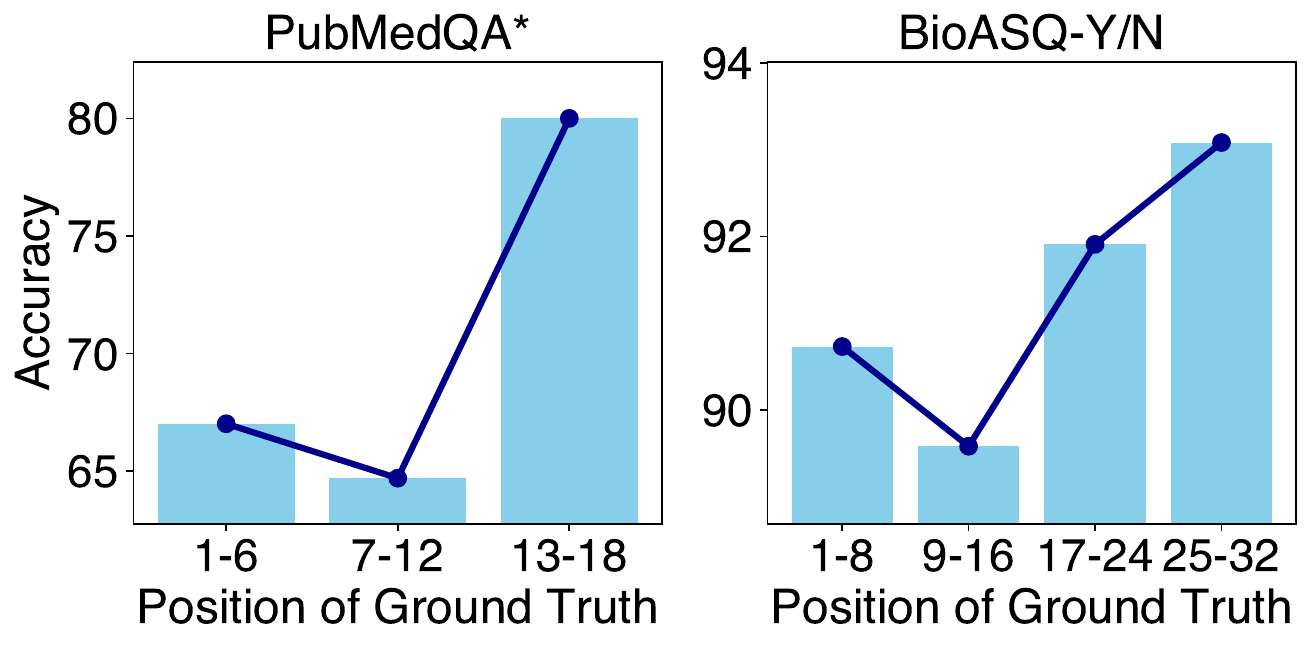}
    \caption{The relations between QA accuracy and the position of the ground-truth snippet in the LLM context.}
    \label{fig:position}
\end{figure}

Figure \ref{fig:position} shows the changes in model accuracy corresponding to different parts of context locations. From the figure, we can see a clear U-shaped decreasing-then-increasing pattern in the accuracy change concerning the position of ground-truth snippets, which sheds light on the arrangement of snippets for medical RAG in future research.

\subsection{Proportion in the MedCorp Corpus}

\begin{figure}[t!]
    \centering
    \includegraphics[width=1\linewidth]{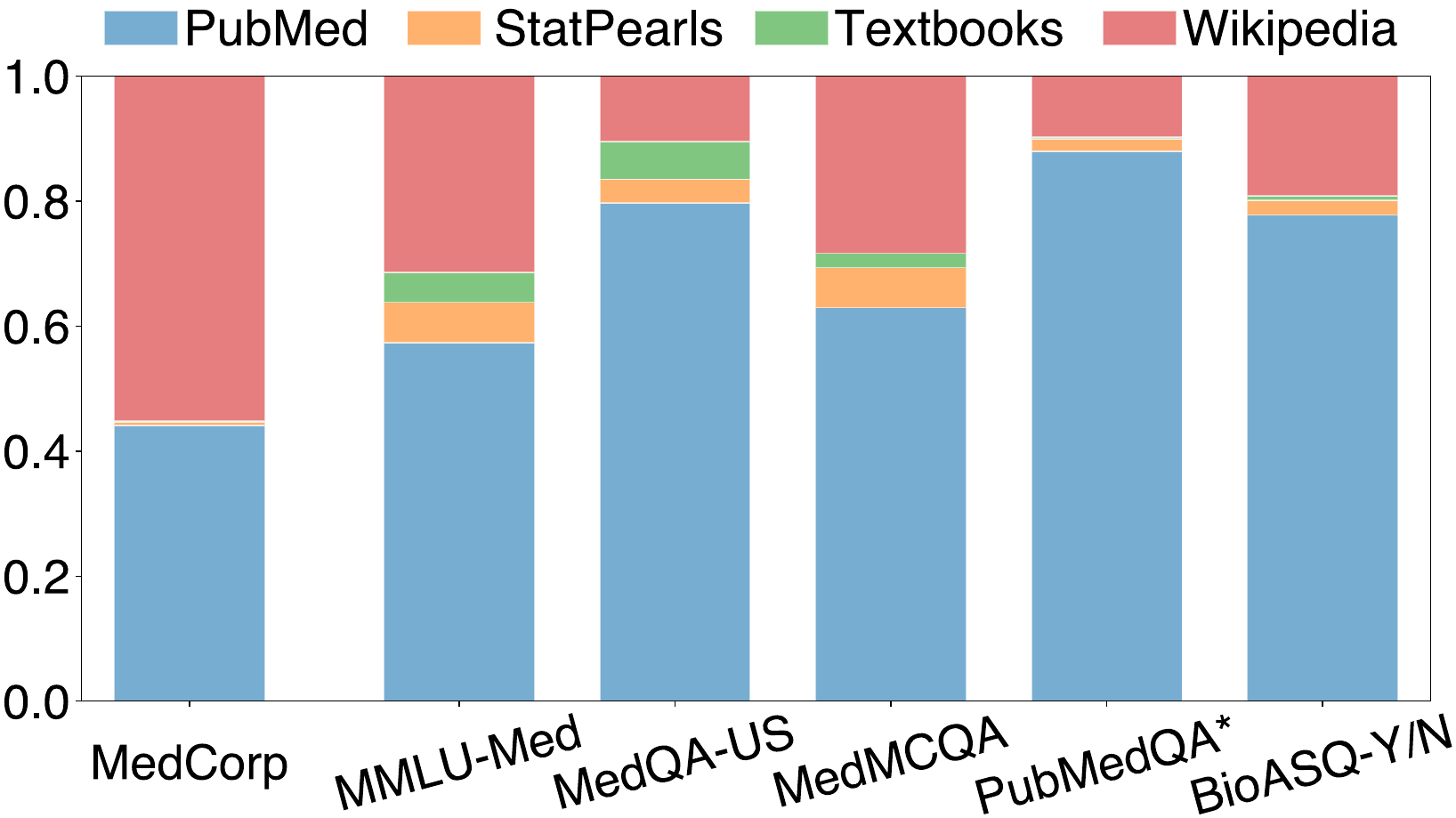}
    \caption{The overall corpus composition of MedCorp and the actually retrieved proportion in different tasks.}
    \label{fig:proportion}
\end{figure}

We also examine the proportion of different sources in the retrieved snippets from MedCorp, and explore how this proportion changes across different tasks. Figure \ref{fig:proportion} displays the proportions of four different sources in MedCorp and the actually retrieved sources in the top 64 retrieved snippets for each task in \textsc{Mirage}. It can be observed from the figure that, in general, the proportion of Wikipedia drops in the retrieved snippets for medical questions, which is expected as many snippets in Wikipedia are not related to biomedicine.

Comparing the distributions for different tasks, there is a task-specific preference pattern. Medical examination tasks (MMLU-Med, MedQA-US, and MedMCQA) tend to have a larger proportion of retrieved snippets from Textbooks and StatPearls. PubMedQA* and BioASQ-Y/N with research-related questions have more relevant snippets from PubMed. The Textbooks corpus has a larger proportion in MedQA-US than in other datasets, which can be explained the fact that this corpus is composed of frequently used textbooks for the US medical licensing examination.

\subsection{Practical Recommendations}

In this section, we discuss the practical indications and recommendations based on our evaluation results of different \textsc{MedRag} settings on \textsc{Mirage}.

\paragraph{Corpus selection.} Results in Table \ref{tab:retrieval_comp} indicate that PubMed and the MedCorp corpus are the only corpora with which \textsc{MedRag} can outperform CoT on all tasks in \textsc{Mirage}. As a large-scale corpus, PubMed serves as a suitable document collection for various kinds of medical questions. If resources permit, the MedCorp corpus could be a more comprehensive and reliable choice: Nearly all \textsc{MedRag} settings using the MedCorp Corpus show improved performance (green-coded cells) compared to the CoT prompting baseline. In general, single corpora other than PubMed are not recommended for medical QA due to their limited volumes of medical knowledge, but they can also be beneficial in specific tasks such as question answering for medical examinations.

\paragraph{Retriever selection.} Among the four individual retrievers used in \textsc{MedRag}, MedCPT is the most reliable one which constantly outperforms other candidates with a higher average score on \textsc{Mirage}. BM25 is a strong retriever as well, which is also supported by other evaluations \citep{thakur2021beir}. The fusion of retrievers can provide robust performance but must be utilized with caution for the retrievers included. As for the PubMed corpus recommended above, a RRF-2 retriever that combines the results from BM25 and MedCPT can be a good selection, since they perform better than the other two with snippets from PubMed. For the MedCorp corpus, both RRF-2 and RRF-4 can be reliable choices, as the corpus can benefit all four individual retrievers in \textsc{MedRag}. 

\paragraph{LLM selection.} Currently, GPT-4 is the best model with about 80$\%$ accuracy on \textsc{Mirage}. However, it is much more expensive than other backbone LLMs. GPT-3.5 can be a more cost-efficient choice than GPT-4, which shows great capabilities of following \textsc{MedRag} instructions. For high-stakes scenarios such as medical diagnoses where patient privacy should be a key concern, the best open-source Mixtral model, which can be deployed locally and run offline, could be a viable option.

%% file: content/conclusion.tex
\section{Conclusion}
To evaluate RAG systems in medicine, we introduced the \textsc{Mirage} benchmark and the \textsc{MedRag} toolkit.
Based on our comprehensive evaluations, we presented many novel observations and practical recommendations to guide the research and real-world deployments of medical RAG systems.

%% file: content/benchmark_appendix.tex
\section{Details of \textsc{Mirage} Datasets}

\paragraph{MMLU-Med.} Massive Multitask Language Understanding (MMLU)\footnote{\url{https://github.com/hendrycks/test}} is a benchmark for the evaluation of the multitask learning capability of language models. The benchmark contains a variety of 57 different tasks \cite{hendrycks2020measuring}. To measure the performance of medical RAG systems, we select a subset of six tasks that are related to biomedicine following \cite{singhal2023large}, including anatomy, clinical knowledge, professional medicine, human genetics, college medicine, and college biology. The subset is collectively denoted as MMLU-Med. Only the test set of each task is used in our benchmark, which contains 1089 questions in total.

\paragraph{MedQA-US.} MedQA\footnote{\url{https://github.com/jind11/MedQA}} \citep{jin2021disease} is a multi-choice QA dataset collected from professional medical board exams. Specifically, we focus on the English part, which includes real-world questions from the US Medical Licensing Examination (MedQA-US). The 1273 four-option test questions are included in our \textsc{Mirage} benchmark.

\paragraph{MedMCQA.} MedMCQA\footnote{\url{https://medmcqa.github.io/}} \citep{pal2022medmcqa} contains 194k multi-choice questions collected from Indian medical entrance exams. The questions cover a wide range of 2.4k healthcare topics and 21 medical subjects. Since the ground truth of its test set is not provided, the dev set of the original MedMCQA is chosen for \textsc{Mirage}, including 4183 medical questions.

\paragraph{PubMedQA*.} PubMedQA\footnote{\url{https://pubmedqa.github.io/}} \citep{jin2019pubmedqa} is a biomedical research QA dataset. It has 1k manually annotated questions constructed from PubMed abstracts. Different from the datasets above, PubMedQA also provides a relevant context for each question to evaluate the reasoning ability of language models. To test the capability of RAG systems to find related documents and answer the question accordingly, we build PubMedQA* by removing given contexts in the 500 expert-annotated test samples of PubMedQA following \cite{lala2023paperqa}. The possible answer to a PubMedQA* question can be yes/no/maybe, reflecting the authenticity of the question statement based on scientific literature.

\paragraph{BioASQ-Y/N.} BioASQ\footnote{\url{http://bioasq.org/}} \citep{tsatsaronis2015overview,krithara2023bioasq} is an annual competition for biomedical QA, which includes both the information retrieval track (Task A) and machine reading comprehension track (Task B). To leverage the resources of BioASQ for our medical RAG benchmark, we select the Yes/No questions in the ground truth test set of Task B from the most recent five years (2019-2023), including 618 questions in total. In the original task, questions are constructed based on biomedical literature, and the ground truth snippets are provided as a basis for machine reading comprehension. Similar to PubMedQA*, BioASQ-Y/N is also a modified version on which RAG systems are supposed to answer the questions without the ground-truth snippet provided.

%% file: content/method_appendix.tex
\section{Detailed Descriptions of \textsc{MedRag}}

\subsection{Document Collections} \label{sec:corpus}

\paragraph{PubMed.} PubMed\footnote{\url{https://pubmed.ncbi.nlm.nih.gov/}} is the most widely used literature resource \citep{lu2011pubmed, jin2024pubmed}, containing over 36 million biomedical articles. Many relevant studies solely use PubMed as the retrieval corpus \citep{frisoni2022bioreader, naik2022literature}. For \textsc{MedRag}, we use a PubMed subset of 23.9 million articles with valid titles and abstracts.

\paragraph{StatPearls.} StatPearls\footnote{\url{https://www.statpearls.com/}} is a point-of-the-care clinical decision support tool similar to UpToDate\footnote{\url{https://www.uptodate.com/}}. We use the 9,330 publicly available StatPearl articles through NCBI Bookshelf\footnote{\url{https://www.ncbi.nlm.nih.gov/books/NBK430685/}} to construct the StatPearls corpus. We chunked StatPearls according to the hierarchical structure, treating each paragraph in an article as a snippet and splicing all the relevant hierarchical headings as the corresponding title. To the best of our knowledge, our work presents the first evaluation of StatPearls in the biomedical NLP community.

\paragraph{Textbooks.} Textbooks\footnote{\url{https://github.com/jind11/MedQA}} \cite{jin2021disease} is a collection of 18 widely used medical textbooks, which are important references for students taking the United States Medical Licensing Examination (USLME). In \textsc{MedRag}, the textbooks are processed as chunks with no more than 1000 characters. We used the $\texttt{RecursiveCharacterTextSplitter}$ from LangChain\footnote{\url{https://www.langchain.com/}} to perform the chunking.

\paragraph{Wikipedia.} As a large-scale open-source encyclopedia, Wikipedia is frequently used as a corpus in information retrieval tasks \cite{thakur2021beir}. We select Wikipedia as one of the corpora to see if the general domain database can be used to improve the ability of medical QA. We downloaded the processed Wikipedia data from HuggingFace\footnote{\url{https://huggingface.co/datasets/wikipedia}} and also chunked the text with LangChain.

\subsection{Retrieval Systems} \label{sec:retriever}

\paragraph{BM25.} BM25 \cite{robertson2009probabilistic} is a commonly used baseline retriever which use bag-of-words and TF-IDF to perform lexical retrieval. In \textsc{MedRag}, BM25 is implemented with Pyserini \cite{lin2021pyserini}\footnote{\url{https://github.com/castorini/pyserini}} using the default hyperparameters to index snippets from all corpora.

\paragraph{Contriever.} Contriever\footnote{\url{https://huggingface.co/facebook/contriever}} \cite{izacard2022unsupervised} is a dense retriever pre-trained on Wikipedia and CCNet \cite{wenzek2020ccnet} with contrastive learning. It is shown to be competitive with BM25 on retrieval tasks in the general domain \cite{thakur2021beir}.

\paragraph{SPECTER.} SPECTER\footnote{\url{https://huggingface.co/allenai/specter}} \cite{cohan2020specter} is a document-level scientific dense retriever which was pre-trained on the Semantic Scholar corpus \cite{ammar2018construction} to encode similar documents with close embeddings.

\paragraph{MedCPT.} MedCPT \citep{jin2023medcpt} is a biomedical embedding model that is contrastively pre-trained by 255 million user clicks from PubMed search logs \citep{fiorini2018user}. It achieved state-of-the-art performance on several biomedical IR tasks. We use the MedCPT Query Encoder\footnote{\url{https://huggingface.co/ncbi/MedCPT-Query-Encoder}} and Article Encoder\footnote{\url{https://huggingface.co/ncbi/MedCPT-Article-Encoder}} to encode the questions and corpus snippets, respectively.

\paragraph{RRF.} \citet{cormack2009reciprocal} proposed to merge results from different retrievers with Reciprocal Rank Fusion (RRF), which effectively fuses the information from different sources by selecting shared predictions. In \textsc{MedRag}, we provide two versions of RRF systems, RRF-2 and RRF-4. RRF-2 is the fusion of results from BM25 and MedCPT, which appear to be the optimal lexical and dense retrievers in our experiments. RRF-4 is a more comprehensive system which fuses the information from all individual retrievers used.

\subsection{Backbone LLMs} \label{sec:llm}

 \paragraph{GPT-3.5 \& GPT-4.} GPT-3.5 and GPT-4 \cite{openai2023gpt4} are two popular commercial LLMs developed by OpenAI, which have already shown great capabilities in answering medical questions \cite{nori2023can,lievin2022can}. In \textsc{MedRag}, we use the specific version of GPT-3.5-turbo-16k-0613 and GPT-4-32k-0613 accessed through Microsoft Azure OpenAI Services\footnote{\url{https://oai.azure.com/}}.
 
 \paragraph{Mixtral.} In \textsc{MedRag}, we use Mixtral-7\(\times\)8B\footnote{\url{https://huggingface.co/mistralai/Mixtral-8x7B-Instruct-v0.1}}, which is an open-source sparse mixture of expert models. Compared with existing open-source models, Mixtral-7\(\times\)8B can achieve both good task performance and fast inference speed \cite{jiang2024mixtral}.
 
 \paragraph{Llama2.} Llama2 \cite{touvron2023llama2} is a series of open-source models that are pre-trained on large-scale data and fine-tuned with human instructions. In \textsc{MedRag}, we use Llama2-70B\footnote{\url{https://huggingface.co/meta-llama/Llama-2-70b-chat-hf}}, which is the largest model in the Llama2 series.

 \paragraph{MEDITRON.} MEDITRON \cite{chen2023meditron} is a series of biomedical LLMs that are built based on Llama2 and fine-tuned on open-source biomedical literature. Its 70B\footnote{\url{https://huggingface.co/epfl-llm/meditron-70b}} version model is contained in \textsc{MedRag}.

 \paragraph{PMC-LLaMA.} PMC-LLaMA \cite{wu2023pmc} is fine-tuned based on LLaMA \cite{touvron2023llama1} using PubMed Central (PMC) papers. Its largest version, PMC-LLaMA-13B\footnote{\url{https://huggingface.co/axiong/PMC_LLaMA_13B}}, is included in \textsc{MedRag}.

%% file: content/appendix.tex
\section{Prompt Templates}

Here are the prompt templates used in our experiments. Figures \ref{fig:prompt_cot} and \ref{fig:prompt_medrag} show the template for all LLMs except MEDITRON. Since the officially released checkpoint of MEDITRON\footnote{\url{https://huggingface.co/epfl-llm/meditron-70b}} is only the pre-trained version without any instruction tuning, it cannot follow the given system prompt well. Therefore, we provide a pseudo one-shot demonstration in the prompt for MEDITRON, where the demonstration does not contain any information of real examples. The templates for MEDITRON are provided in Figures \ref{fig:prompt_cot_meditron} and \ref{fig:prompt_medrag_meditron}.

\begin{figure*}[h]
\begin{AIbox}{Prompt template for medical QA with CoT}
You are a helpful medical expert, and your task is to answer a multi-choice medical question. Please first think step-by-step and then choose the answer from the provided options. Organize your output in a json formatted as Dict\{``step\_by\_step\_thinking'': Str(explanation), ``answer\_choice'': Str\{A/B/C/...\}\}. Your responses will be used for research purposes only, so please have a definite answer. \\

Here is the question: \\
\verb|{{question}}|\\

Here are the potential choices:\\
\verb|{{options}}|\\

Please think step-by-step and generate your output in json:
\end{AIbox}
\caption{Template used to generate prompts for medical QA with CoT.}
\label{fig:prompt_cot}
\end{figure*}

\begin{figure*}[h]
\begin{AIbox}{Prompt template for medical QA with \textsc{MedRag}}
You are a helpful medical expert, and your task is to answer a multi-choice medical question using the relevant documents. Please first think step-by-step and then choose the answer from the provided options. Organize your output in a json formatted as Dict\{"step\_by\_step\_thinking": Str(explanation), "answer\_choice": Str\{A/B/C/...\}\}. Your responses will be used for research purposes only, so please have a definite answer. \\

Here are the relevant documents:\\
\verb|{{context}}|\\

Here is the question: \\
\verb|{{question}}|\\

Here are the potential choices:\\
\verb|{{options}}|\\

Please think step-by-step and generate your output in json:
\end{AIbox}
\caption{Template used to generate prompts for medical QA with \textsc{MedRag}.}
\label{fig:prompt_medrag}
\end{figure*}

\begin{figure*}[h]
\begin{AIbox}{Prompt template for medical QA with CoT on MEDITRON}
You are a helpful medical expert, and your task is to answer a multi-choice medical question. Please first think step-by-step and then choose the answer from the provided options. Organize your output in a json formatted as Dict\{``step\_by\_step\_thinking'': Str(explanation), ``answer\_choice'': Str\{A/B/C/...\}\}. Your responses will be used for research purposes only, so please have a definite answer. \\

\#\#\# User:
Here is the question:\\
...\\

Here are the potential choices:\\
A. ...\\
B. ...\\
C. ...\\
D. ...\\
X. ...\\

Please think step-by-step and generate your output in json.\\

\#\#\# Assistant:\\
\{``step\_by\_step\_thinking'': ..., ``answer\_choice'': ``X''\}\\

\#\#\# User:\\
Here is the question:\\
\verb|{{question}}|\\

Here are the potential choices:\\
\verb|{{options}}|\\

Please think step-by-step and generate your output in json.\\

\#\#\# Assistant:
\end{AIbox}
\caption{Template used to generate prompts for medical QA with CoT on MEDITRON.}
\label{fig:prompt_cot_meditron}
\end{figure*}

\begin{figure*}[h]
\begin{AIbox}{Prompt template for medical QA with \textsc{MedRag} on MEDITRON}
You are a helpful medical expert, and your task is to answer a multi-choice medical question using the relevant documents. Please first think step-by-step and then choose the answer from the provided options. Organize your output in a json formatted as Dict\{"step\_by\_step\_thinking": Str(explanation), "answer\_choice": Str\{A/B/C/...\}\}. Your responses will be used for research purposes only, so please have a definite answer. \\

Here are the relevant documents:\\
\verb|{{context}}|\\

\#\#\# User:\\
Here is the question:\\
...\\

Here are the potential choices:\\
A. ...\\
B. ...\\
C. ...\\
D. ...\\
X. ...\\

Please think step-by-step and generate your output in json.\\

\#\#\# Assistant:\\
\{``step\_by\_step\_thinking'': ..., ``answer\_choice'': ``X''\}\\

\#\#\# User:\\
Here is the question:\\
\verb|{{question}}|\\

Here are the potential choices:\\
\verb|{{options}}|\\

Please think step-by-step and generate your output in json.\\

\#\#\# Assistant:
\end{AIbox}
\caption{Template used to generate prompts for medical QA with \textsc{MedRag} on MEDITRON.}
\label{fig:prompt_medrag_meditron}
\end{figure*}